\documentclass[letterpaper, 10 pt, conference]{ieeeconf}  % Comment this line out if you need a4paper

\IEEEoverridecommandlockouts                              % This command is only needed if 
\usepackage{amsmath} % assumes amsmath package installed
\usepackage{amssymb}  % assumes amsmath package installed
\usepackage{dsfont}
\usepackage[utf8x]{inputenc}
\usepackage{graphicx}
\usepackage{atbegshi,picture}

\renewcommand{\vec}[1]{{\boldsymbol{{#1}}}} % vector
\newcommand{\mat}[1]{{\boldsymbol{{#1}}}} % matrix

\title{\LARGE \bf
CNNATT: Deep EEG \& fNIRS Real-Time Decoding of bimanual forces}

\author{Pablo Ortega$^{1}$, Tong Zhao$^{2}$ and Aldo Faisal$^{1,2}$% <-this % stops a space
\thanks{Brain \& Behaviour Lab, $^{1}$Dept. of Computing,
        $^{2}$Dept. of Bioengineering, Imperial College London,  
        London SW7 2AZ, UK; Correspondence: aldo.faisal@imperial.ac.uk.We acknowledge EPSRC's CDT HiPEDS (EP/L016796/1) and their  capital equipment grant, as well as UKRI's Turing AI Fellowship to AAF.
        }%
}

\AtBeginShipoutNext{\AtBeginShipoutUpperLeft{%
  \put(\dimexpr\paperwidth-1cm\relax,-1.5cm){\makebox[0pt][r]{Accepted manuscript IEEE/EMBS Neural Engineering (NER) 2021}}%
}}

\begin{document}

\maketitle
\thispagestyle{empty}
\pagestyle{empty}

%%%%%%%%%%%%%%%%%%%%%%%%%%%%%%%%%%%%%%%%%%%%%%%%%%%%%%%%%%%%%%%%%%%%%%%%%%%%%%%%
\begin{abstract}
 Non-invasive cortical neural interfaces have only achieved modest performance in cortical decoding of limb movements and their forces, compared to invasive brain-computer interfaces (BCIs). 
 While non-invasive methodologies are safer, cheaper and vastly more accessible technologies, signals suffer from either poor resolution in the space domain (EEG) or the temporal domain (BOLD signal of functional Near Infrared Spectroscopy, fNIRS). 
 The non-invasive BCI decoding of bimanual force generation and the continuous force signal has not been realised before and so we introduce an isometric grip force tracking task to evaluate the decoding. 
 We find that combining EEG and fNIRS using deep neural networks works better than linear models to decode continuous grip force modulations produced by the left and the right hand.
 Our multi-modal deep learning decoder achieves 55.2 FVAF[\%] in force reconstruction and improves the decoding performance by
 at least 15\% over each individual modality. 
 Our results show a way to achieve continuous hand force decoding using cortical signals obtained with non-invasive mobile brain imaging has immediate impact for rehabilitation, restoration and consumer applications.
\end{abstract}

%%%%%%%%%%%%%%%%%%%%%%%%%%%%%%%%%%%%%%%%%%%%%%%%%%%%%%%%%%%%%%%%%%%%%%%%%%%%%%%%
\section{INTRODUCTION}

Brain computer interfaces (BCIs) offer an alternative way to interact with our environment.
They are specially relevant for people whose natural ability to physically interact has been lost or damaged. 
There has been a lot of progress in the decoding of kinematic variables in {BCI} \cite{carmena2003learning,bansal2012decoding,naufel2019muscle}.
In contrast, human force decoding is less explored even though force is essential for safe and meaningful mechanical interactions \cite{xiloyannis2017gaussian}.
In addition, force decoding can provide more generalisable {BCI} decoders \cite{cherian2011motor} specially in changing dynamic conditions \cite{chhatbar2013towards}.
Even in the cases where the ultimate goal of a {BCI} is different from force decoding, understanding and exploiting signals that already exist in the brain might offer a more intuitive control than learning new ones \cite{flint2014extracting}.
Force control is specially relevant for the hands. 
As the specie with greatest hand dexterity we spend most of our day physically interacting with our environment through our hands.
Indeed, for completely paralysed people the recovery of the hand function is the most relevant priority \cite{anderson2004targeting}.

To enable intuitive {BCI} hand force control we first need to understand how to decode force from cortical signals in healthy humans.
Invasive {BCI} have done a lot of progress in the force decoding field in humans \cite{ganzer2020restoring} and monkeys \cite{carmena2003learning}.
More recently, invasive BCIs showed the advantages of non-linear approaches in force decoding \cite{naufel2019muscle}.
However they have a low acceptance rate across potential users due to the several surgical procedures they require. In contrast, for non-invasive {BCI}, force decoding still remains relatively unexplored.
Discrete force characteristics have been decoded from multi-modal ({EEG} and {fNIRS}) signals.
In \cite{hansen2014improved} two force levels ($20\%$ and $60\%$ of the maximum voluntary contraction, MVC) were decoded and force detection was performed for the right foot.
In \cite{yin2015hybrid} and \cite{fu2016imagined} a similar multi-modal approach was used for the classification of force imageries.
These studies showed that the combined use of {fNIRS} and {EEG} provided a significant advantage in classification of discrete force characteristics.
In \cite{paek2019regression} {EEG} was used to decode unimanual force trajectories and reported a modest reconstruction performance ($\text{correlation}\approx0.42$). 
However, the multi-modal ({EEG} and {fNIRS}) approach has not yet been explored for the decoding of continuous force trajectories. 
Furthermore, to the best of our knowledge, no non-invasive BCI study has explored the simultaneous production of force with the right and the left hand despite bimanual interactions being more frequent in daily activities.

We use here a multi-modal system ({fNIRS} and {EEG}) to continuously decode bimanual force trajectories and explore the advantages that deep learning (DL) introduces in the fusion of signals with different neurophysiological origins.

\section{METHODS AND MATERIALS}

\begin{figure}
    \centering
    \includegraphics[width=0.4\textwidth]{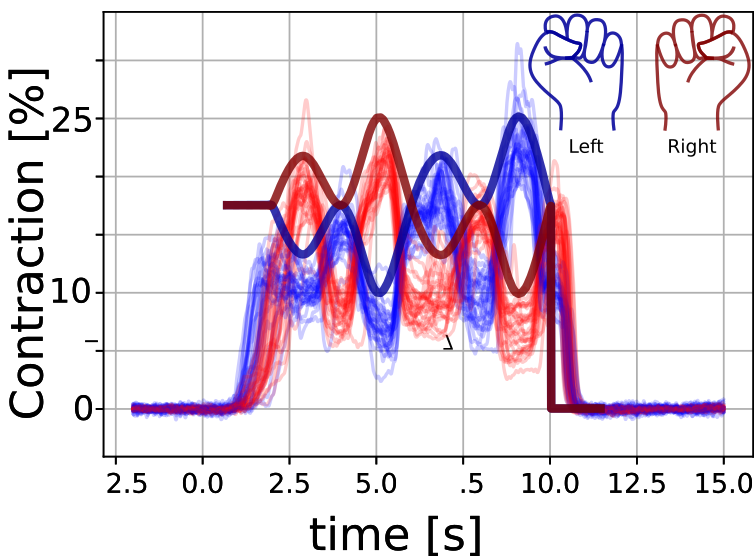}
    \caption{fNIRS and EEG recording setup during bimanual force generation. Aggregated contraction trials for one of the conditions.}
    \label{fig:setup}
\end{figure}

\subsection{Protocol and task}
Ten participants ($N=10$) were asked to perform a bimanual isometric contraction task.
We provided the force profile that each subject had to track with each hand with two characteristics (Fig. \ref{fig:setup}).
First, both hands were either contracted or relaxed at the same time (relaxation vs contraction). 
Second, in the contraction state the hands had to dynamically track four force profiles with different crest orders as that presented in Fig. \ref{fig:setup}.
The dynamic force that each hand had to track was different and introduced contraction variability during the contraction state.
The different way each hand was engaged during the dynamic force tracking enables a better representation of continuous control of force that corresponds to more natural bimanual manipulations. 

The participants received visual feedback on the desired contraction trajectory they had to follow with each hand.
Four conditions (one condition per force profile) were used. 
Each condition represented a different force trajectory which increased the variability of the brain signals and contractions recorded. 
All conditions lasted $10\,$s and all participants did $30$ trials per condition.
The order of the conditions was randomised but each condition was performed in blocks of $30$ trials. 
The highest level of contraction was set to $25\%\,$MVC and the lowest to $10\%\,$MVC.
The trajectories for both hands were designed so that the average of the contraction during the $10\,$s corresponded to $17.5\%~$MVC.
Each trial was followed by a randomised resting period uniformly distributed between $15$ and $21\,$seconds, to avoid phasic constructive interference of systemic artefacts, e.g. Mayer waves, in the brain responses.
The refreshing of the feedback in the screen was set to $100~$Hz.

Participants were right-handed (confirmed by the Edinburgh inventory).
The Imperial College Research Ethics Committee approved all procedures and all participants gave their written informed consent.
The experiment complied with the Declaration of Helsinki for human experimentation and national and applicable international data protection rules.

\subsection{Recordings}

Twenty four ($N=24$) co-aligned {EEG} and {fNIRS} channels covered the bilateral sensorimotor cortex and were used as the brain signals from which to decode the force signals generated with each hand (Fig. \ref{fig:setup}). 

%%% fNIRS was recorded
The {fNIRS} signals were recorded using a NIRScout system (NIRx Medizintechnik GmbH, Berlin, Germany). 
We used a total of 12 optodes per hemisphere (10 sources and 8 detectors in total) sampling at $12.5~$Hz.

%%% EEG was recorded 
{EEG} was recorded using $24$ channels of an ActiChamp amplifier (BrainProducts, Berlin, Germany) operating at $4~$kHz (running software BrainVision, v1.20.0801).
{EEG} was first downsampled to $250~$Hz (with anti-aliasing down-pass filtering).
Notch filters were applied at the mains ($50~$Hz) and {fNIRS} ($12.5~$Hz) frequencies and their harmonics.
{EEG} was finally high-pass filtered above $1~$Hz using a $5$th order Butterworth filter.
ICA was then applied to automatically remove {EOG} artefacts rejecting $1$ component when they had a correlation above $0.3$ in absolute value.

%%% Force was recorded
Two grip force transducers (PowerLab 4/25T, ADInstruments, Castle Hill, Australia) were used to record the force generated by each hand simultaneously recording at $1\,$kHz. 
The force signals were first resampled to $250\,$Hz, then band-pass filtered between $0.01$ and $10\,$Hz with a Butterworth filter of order 3 and then again high-pass filtered with an elliptical filter of order 1 above $0.01\,$Hz.
Drift was further eliminated removing the linear drift per trial.
Force measures were finally converted to contraction values using the recorded {MVC} before the experiment started.

\subsection{Preprocessing}
All channels were used in the analysis and needed preprocessing.
Optical intensities were low-pass filtered below $0.25\,$Hz with a $7$th order elliptical filter.
Changes in optical densities per wavelength, $\Delta \mbox{OD} ^\lambda_{ij}(t)$, were obtained using Beer-Lambert's law.

{EEG} was used to extract the power and phases in different {EEG} bands using the Hilbert transform.%\cite{oppenheim2009adaptive}.
The following frequency bands were used: delta ($1-4\,$Hz), theta ($4-8\,$Hz), alpha ($8-13\,$Hz), beta ($13-30\,$Hz), low-gamma ($30-50\,$Hz), mid-gamma ($70-110\,$Hz) and high-gamma ($130-200$). 
Finally, these EEG Hilbert features were resampled to $12.5\,$Hz to have the same sampling frequency than the fNIRS signals.

We additionally recorded the hemodynamic activity of the scalp skin on the forehead using a NONIN 8000R (Tilburg, The Netherlands).
The skin hemodynamics reflect the variations of hemoglobin due to the heart and breathing activity but do not contain brain hemodynamic responses. 
We used scalp hemodynamics to discard that pulse and breathing were predictive of force.
All epochs were extracted from $4\,$s before the ``Go'' instruction to $14\,$s after.

\subsection{Decoding methods}

\begin{figure}
    \centering
    \vspace{0.25cm}
    \includegraphics[width=0.49\textwidth]{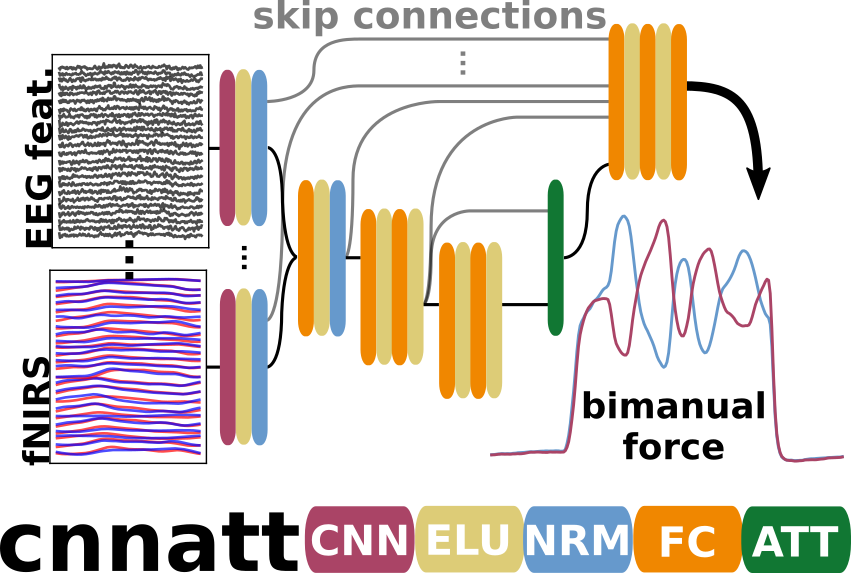}
    \caption{Architecture of \emph{cnnatt}. Network layers are denoted by vertical rounded bars, colour coded as follows exponential-linear units (ELU), normalisation (NRM), fully connected (FC) and self-attention layers (ATT).}
    \label{fig:arch}
\end{figure}

The {fNIRS}, the EEG Hilbert features and force are resampled so all measures could be aligned in time.
EEG Hilbert features and fNIRS are then used to build a linear and a deep learning (cnnatt) model to decode the bimanual force. 
Both decoders used $800\,$ms of brain signals (EEG Hilbert features and fNIRS signal) history to decode the bimanual force.
% Although most {BCI} decoders exploiting electrical signals use memory depths below $500\,$ms we increased the memory to $800\,$ms to better capture the slower {fNIRS} signals without excessively increasing the input dimensionality.

\begin{figure}[t]
    \centering
    \vspace{0.4cm}
    \includegraphics[width=0.45\textwidth]{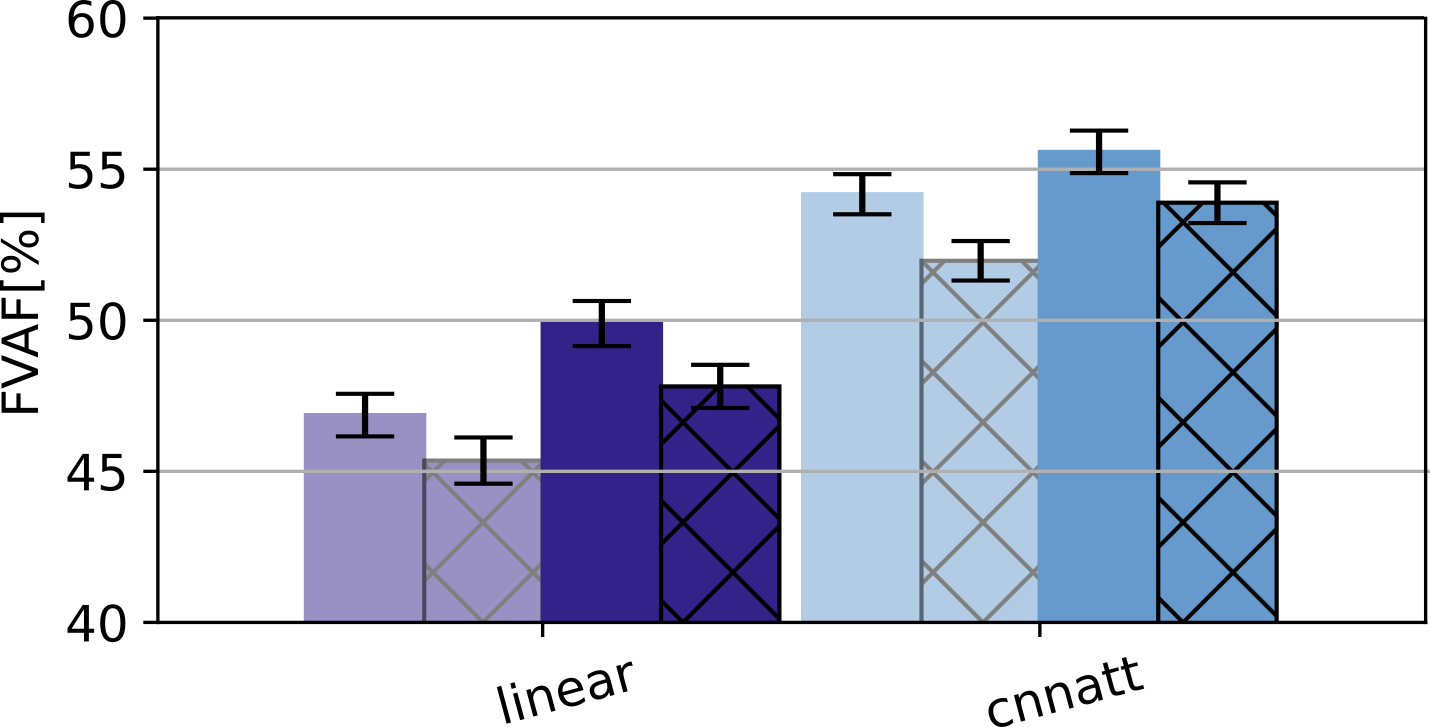}
    \caption{
    Fraction of variance accounted for (FVAF) by the linear and \emph{cnnatt} force predictions using multi-modal (EEG \& fNIRS) signals.
    Darker bars correspond to the right hand and brighter ones to the left hand.
    Patterned bars correspond to the FVAF[\%] between the decoded hand and the opposite real hand to test for specificity of hand decoding. 
    The \emph{cnnatt} model achieves a better performance compared to the linear model. 
    }
    \label{fig:FVAF}
\end{figure}
The decoding can be expressed as 
$\vec f_t =  \phi (\mat X_{t-800\text{ms},\, ...,\, t} )$
where $\vec f_t$ represents the vector of bimanual left (L) and right (R) forces $\vec f_t = \left[ f_{\text{L},t}, f_{\text{R},t} \right]^{\text{T}}$ at time $t$, and $\mat X_{t-800\text{ms}, ..., t}$ the matrix of {fNIRS} and {EEG} features from $t-800\text{ms}$ to $t$.

The linear decoder was trained using the Lasso method.  Our deep learning \emph{cnnatt} model including CNN and attention layers (Fig. \ref{fig:arch}), was trained using the mean squared error loss, Adam optimiser and early stop techniques.

\section{RESULTS}

\begin{figure}[t]
    \centering
    \vspace{0.3cm}
    \includegraphics[width=0.45\textwidth]{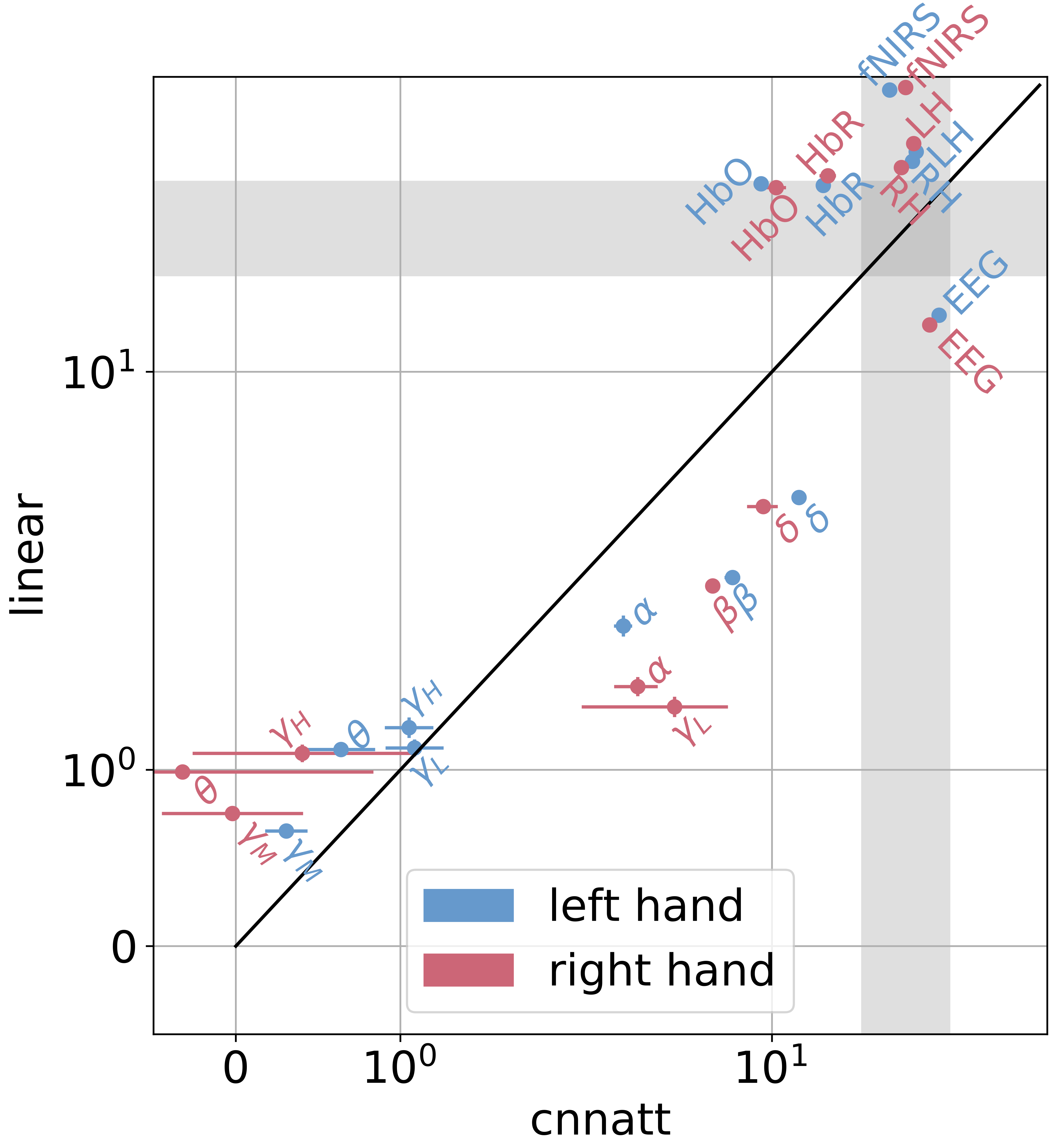}
    \caption{
    Comparison of sensitivity of the linear and cnnatt decoding to the perturbation of the time order of the signal. 
    The linear decoding is more dependent on fNIRS than EEG (outside the horizontal grey bar) while the cnnatt model has a more balanced dependence on both (inside vertical grey bar). 
    }
    \label{fig:sensitivity}
\end{figure}

We evaluate the force reconstruction performance of our multi-modal decoding approach using the fraction of variance accounted for
$\mbox{FVAF[\%]} = 100 \cdot ( 1 - {\sum_{i=1}^N(y_i-\hat{y}_i)^2}/{\sum_{i=1}^N(y_i-\bar{y})^2} )$.
Here, $y_i$ is the i\textsuperscript{th} sample of the real signal, $\hat{y}_i$ is the corresponding sample of the predicted signal, $\bar{y}$ is the average of the real signal and $N$ the number of samples in the signal.
The fraction of variance accounted for (FVAF[\%]) has a value between $(-\inf, 100\%]$.
A $100\%\,$FVAF represents a total reconstruction.
A $0\%$ represents a reconstruction that is as good as using the average of the signal as predictor and negative FVAF[\%] even worse reconstructions. 
We use EEG and fNIRS signals to decode bimanual force trajectories.

At the population level, the increase of reconstruction performance due to the combination of {fNIRS} and {EEG} (multi-modal, $48.5\,$FVAF[\%]) is significant when compared to either {EEG} ($33.5\,$FVAF[\%]) or {fNIRS} ($40\,$FVAF[\%]) alone and achieves a maximum $15\%$ increase in FVAF[\%] for the multi-modal signals (Kruskal Wallis, Tukey corrected comparison on FVAF[\%],  $p<0.05$). 

%%% First, to see if DL is better than linear we compare FVAF... 
First, we compare the capabilities of a linear and deep learning model to reconstruct force in a real-time and causal way. 
As shown in figure \ref{fig:FVAF}, for both hands cnnatt achieves a better force reconstruction (left/right $54.2/55.6\,$FVAF[\%]) than the linear model (left/right $46.9/49.9\,$FVAF[\%], Kruskal-Wallis, Tukey corrected comparison, $p<0.05$). 
The crossed bars in the figure indicate the performance when the decoded hand is used as reconstruction of the opposite hand to test for hand decoding specificity. 
For both models, this leads to lower FVAF[\%] suggesting that the decoding is specific of the hand (linear $45.4/47.8$, cnnatt $52.0/54.0\,$FVAF[\%], only significant for the right hand of cnnatt, Kruskal-Wallis test, $p<0.05$).  

Second, to understand the dependency of each decoding approach (linear or cnnatt) to each of the multi-modal input features we perform a sensitivity analysis. 
Figure \ref{fig:sensitivity} shows the comparison of each model sensitivity to the perturbation of each input feature. 
The sensitivity is computed using a perturbation approach in which we randomly shuffle the time dimension of the input feature we want to analyse and measure its impact in the force decoding FVAF[\%]. 
This test evaluates how important the temporal evolution of the features and their auto-correlation are for the decoding (in contrast to their amplitude distribution).
To standardise this measure we compute the percent change in performance with a $0\%$ change representing the performance of the unperturbed signals and $100\%$ the maximum performance reduction when all features are perturbed. 
Namely, the higher the sensitivity to an input feature perturbation, the more dependant the model is on that feature for an accurate decoding.

As we can see in Figure \ref{fig:sensitivity}, both models are sensitive to perturbations of any of the features (positive changes) but never to the same extent ($100\%$ level) as when all features are perturbed simultaneously.
This shows that the temporal structure of the signal is important in the decoding and suggests that the decoding has a causal nature.

The comparison of the linear and the DL system shows that the latter strikes a better balance in its dependency between EEG and fNIRS signals, both inside the grey vertical bar (Fig. \ref{fig:sensitivity}, multiple comparison of means, Tukey correction, $p>0.01$ for the right hand and $p<0.01$ for the left hand).
In contrast, for a same dependency range (horizontal grey bar) the linear decoder has a much higher dependency in fNIRS than in EEG (multiple comparison of means, Tukey correction, $p<0.01$ for left and right hand).
Both systems can be applied in real-time but cnnatt is slower ($21.0\pm0.0\,$ms per trial) than the linear model ($2\pm0.5\,$ms).
Finally, we verify that breathing or pulse rhythms do not have predictive power on force generation.
Namely, when skin hemodynamics are used to decode force with a similar linear model trained with this purpose, they have a significant lower reconstruction performance ($\mbox{FVAF[\%]}=1\%$) compared to when fNIRS is used ($\mbox{FVAF[\%]}=32.5\%$, t-test, $\mbox{p}<0.001$).

\section{DISCUSSION}

We set us the challenge of reconstructing continous bimanual grip force production in a dynamic task. We tackled the challenge by using multi-modal non-invasive signals (EEG \& fNIRS). 
We introduce a bimanual isometric task that opens up new BCI challenge. Using a linear model we first show that the fusion of mul~-modal signals brings a   $15\,$FVAF[\%] increase compared to only using one of the two signals. 
To explore the breadth of non-linear models we crafted a deep learning architecture (\emph{cnnatt}). We show that our \emph{cnnatt} deep learning model improves the bimanual force reconstruction in terms of FVAF[\%] in $5$ to $8$ points compared to the linear system (Fig.~\ref{fig:FVAF}).
Both models preserve the specificity to the decoded hand as shown by the decay of reconstruction performance when the FVAF[\%] is computed in the opposite hand than the one the model was trained to decode.

Our sensitivity results show that the multi-modal linear decoder is more dependent on fNIRS than on EEG while DL  better exploits all EEG bands (Fig. \ref{fig:sensitivity}).
In particular, in combination with fNIRS, DL achieves a better exploitation of the delta and beta bands ($p<0.05$, Kruskal-Wallis test, Tukey correction) which supports \cite{paek2019regression} and expands their results to the bimanual multi-modal case.
We note that the distribution of fNIRS amplitudes is centred and has a higher standard deviation, while the EEG Hilbert features have a skewed and narrower distribution and the target force distribution is bimodal - which  convolutional layers appear to capture \emph{en passant}.
% We leave the interpretation of the features learned by the model for future work.

\section{CONCLUSION}
We used Deep Learning to solve the  multi-modal sensor fusion and decoding problem and were able to decode continuous  force generated in a dynamic bi-manual grip force task. Combining EEG and fNIRS is particularly challenging as signals differ by 3 orders of magnitude in time scale (ms vs s), thus we used the power of representational learning in deep learning. Previous approaches used the advantages of Gaussian Process regression to achieve efficient continuous multi-modal decoding \cite{xiloyannis2017gaussian}, however signals there the EEG and MMG signals operated on similar timescales.
Deep Learning is  data-hungry and thus a challenge for BCI, however our approach can be directly mapped to data-efficiency improving meta-learning \cite{li2021meta} and multi-subject transfer learning \cite{wei2021transfer} recently demonstrated in Deep EEG BCI.  
We show that non-invasive human interfacing can overcome continuous decoding challenges usually thought the realm of invasive BCI. Combining EEG-fNIRS has direct implications in BCI for restoration of movement and robotic control \cite{dziemian2016gaze}, but also for real-world and consumer use \cite{haar2020brain}.

%\addtolength{\textheight}{-12cm}   % This command serves to balance the column lengths
                                  % on the last page of the document manually. It shortens
                                  % the textheight of the last page by a suitable amount.
                                  % This command does not take effect until the next page
                                  % so it should come on the page before the last. Make
                                  % sure that you do not shorten the textheight too much.

%%%%%%%%%%%%%%%%%%%%%%%%%%%%%%%%%%%%%%%%%%%%%%%%%%%%%%%%%%%%%%%%%%%%%%%%%%%%%%%%

%%%%%%%%%%%%%%%%%%%%%%%%%%%%%%%%%%%%%%%%%%%%%%%%%%%%%%%%%%%%%%%%%%%%%%%%%%%%%%%%

%%%%%%%%%%%%%%%%%%%%%%%%%%%%%%%%%%%%%%%%%%%%%%%%%%%%%%%%%%%%%%%%%%%%%%%%%%%%%%%%
% \section*{APPENDIX}

% Appendixes should appear before the acknowledgment.

%%%%%%%%%%%%%%%%%%%%%%%%%%%%%%%%%%%%%%%%%%%%%%%%%%%%%%%%%%%%%%%%%%%%%%%%%%%%%%%%

\bibliographystyle{ieeetrans}
\bibliography{root.bib}

\end{document}